\newif\ifarxiv 
\arxivfalse

\newif\ifreview 
\reviewtrue

\newif\iffinal 
\finaltrue

\ifarxiv
\documentclass{article}
\else
\ifreview
\documentclass[dvipsnames,format=sigconf,anonymous=false,review=false]{acmart}
\else
\documentclass[dvipsnames,format=sigconf]{acmart}
\fi
\fi

\usepackage[utf8]{inputenc}

\ifarxiv
\usepackage[a4paper,left=2cm,right=2cm,top=2cm,bottom=2cm]{geometry}
\usepackage[onehalfspacing]{setspace}
\usepackage[numbers,longnamesfirst]{natbib}
\usepackage{amsthm,amsthm,amssymb}
\usepackage{authblk}
\usepackage[hidelinks]{hyperref}
\fi

\usepackage{amsmath,mathtools} 
\usepackage{dsfont}
\usepackage{makecell}

\usepackage[vlined, linesnumbered, ruled]{algorithm2e}
\SetKwComment{Comment}{/* }{*/}
\SetKwInput{KwData}{Input}
\SetKwInput{KwResult}{Output}
\IncMargin{1.5em}

\usepackage{booktabs} 
\usepackage{xspace}
\usepackage{graphicx}
\usepackage{xspace}
\usepackage{url}
\usepackage{enumitem}
\setlist[enumerate,1]{label=\textup{(\roman*)}}

\usepackage{caption} 
\usepackage{subcaption}

\usepackage{booktabs}
\usepackage{svg}

\SetAlgoSkip{}
\DontPrintSemicolon

\usepackage[appendix=append]{apxproof}  

\allowdisplaybreaks[3] 

\usepackage{tikz}
\usetikzlibrary{shapes,arrows}
\usetikzlibrary{decorations}
\usetikzlibrary{decorations.markings}
\usetikzlibrary{plotmarks}
\usetikzlibrary{calc}
\usetikzlibrary{mindmap}
\usetikzlibrary{shadows}
\usetikzlibrary{backgrounds}
\usetikzlibrary{patterns}
\usetikzlibrary{shapes.symbols}

\usepackage{pgfplots}
\usepackage{pgfplotstable}
\usepgfplotslibrary{statistics}
\pgfplotsset{compat=1.18}

\ifarxiv
\newtheorem{theorem}             {Theorem}

\newtheorem{definition} [theorem]{Definition}

\else
\newtheoremrep{theorem}{Theorem}
\newtheoremrep{lemma}     [theorem]{Lemma}
\newtheoremrep{definition}[theorem]{Definition}
\newtheoremrep{conjecture}[theorem]{Conjecture}
\newtheoremrep{corollary} [theorem]{Corollary}
\newtheoremrep{proposition}[theorem]{Proposition}
\fi

\usepackage{tabularx,colortbl}

\makeatletter
\g@addto@macro\bfseries{\boldmath}
\makeatother

\DeclareFixedFont{\ttb}{T1}{txtt}{bx}{n}{12} 
\DeclareFixedFont{\ttm}{T1}{txtt}{m}{n}{12}  

\usepackage{color}
\definecolor{deepblue}{rgb}{0,0,0.5}
\definecolor{deepred}{rgb}{0.6,0,0}
\definecolor{deepgreen}{rgb}{0,0.5,0}

\usepackage{listings}

\newcommand\pythonstyle{\lstset{
language=Python,
basicstyle=\ttm,
morekeywords={self},              
keywordstyle=\ttb\color{deepblue},
emph={MyClass,__init__},          
emphstyle=\ttb\color{deepred},    
stringstyle=\color{deepgreen},
frame=tb,                         
showstringspaces=false
}}

\lstnewenvironment{python}[1][]
{
\pythonstyle
\lstset{#1}
}
{}

\newcommand\pythoninline[1]{{\pythonstyle\lstinline!#1!}}

\ifarxiv\else 
\fi

\newcommand{\toggleplot}[1]{{\textcolor{red}{Plots removed to increase compilation speed. Use command $\backslash$toggleplot in preamble to reinsert them.}}}

\iffinal
\newcommand{\andre}[1]{}
\newcommand{\romann}[1]{}
\newcommand{\cuong}[1]{}
\newcommand{\newedit}[1]{}
\else
\newcommand{\andre}[1]{\textcolor{green!50!black}{[(Andre) #1]}}
\newcommand{\romann}[1]{\textcolor{purple}{[(Roman) #1]}}
\newcommand{\cuong}[1]{\textcolor{orange}{[(Cuong) #1]}}
\newcommand{\newedit}[1]{\textcolor{blue}{#1}}

\fi

\ifarxiv
\else
\copyrightyear{2026}
\acmYear{2026}
\acmConference[GECCO Companion '26]{Genetic and Evolutionary Computation Conference}{July 13--17, 2026}{San Jose, Costa Rica}
\acmBooktitle{Genetic and Evolutionary Computation Conference (GECCO Companion '26), July 13--17, 2026, San Jose, Costa Rica}
\acmDOI{10.1145/3795101.3814688}
\acmISBN{979-8-4007-2488-6/2026/07}
\fi

\ifarxiv
\else

\begin{document}

\ifarxiv
\author[1]{Duy Long Tran}
\author[2]{Anja Jankovic}
\author[3]{Marie Anastacio}
\author[4]{Holger Hoos}
\author[5]{Roman~Kalkreuth}
\affil[1]{Chair for AI Methodology, Faculty of Computer Science, RWTH~Aachen University, Aachen, Germany}
\date{}
\else
\author{Duy Long Tran}
\orcid{0009-0002-0790-1410}
\affiliation{
  \institution{Chair for AI Methodology, RWTH~Aachen University}
  \city{Aachen}
  \country{Germany}
}
\email{duy.long.tran@rwth-aachen.de}

\author{Anja Jankovic}
\orcid{0000-0001-9267-4595}
\affiliation{
  \institution{Chair for AI Methodology, RWTH~Aachen University}
  \city{Aachen}
  \country{Germany}
}
\email{jankovic@aim.rwth-aachen.de}
  
\author{Marie Anastacio}
\orcid{0000-0002-4039-2470}
\affiliation{
  \institution{Chair for AI Methodology, RWTH~Aachen University}
  \city{Aachen}
  \country{Germany}
}
\email{anastacio@aim.rwth-aachen.de}

\author{Holger Hoos}
\orcid{0000-0003-0629-0099}
\affiliation{
  \institution{Chair for AI Methodology, RWTH~Aachen University}
  \city{Aachen}
  \country{Germany}
}
\email{hh@aim.rwth-aachen.de}

\author{Roman~Kalkreuth}
\orcid{0000-0003-1449-5131}
\affiliation{
  \institution{Chair for AI Methodology, RWTH~Aachen University}
  \city{Aachen}
  \country{Germany}
}
\email{kalkreuth@aim.rwth-aachen.de}

\fi

\title{Improving Evaluation of Recombination-based \newline{} Cartesian Genetic Programming}

\ifarxiv
\maketitle
\fi

\begin{abstract}
Cartesian Genetic Programming has traditionally been using mutation as its main and often sole genetic operator to drive evolutionary search. Despite advancements in recent years, 
recombination-based approaches have long been avoided, 
due to apparent lack of performance gains. This study examines two recently suggested recombination-based operators, subgraph crossover and discrete phenotypic recombination on SRBench, a benchmarking platform for symbolic regression. Using the implementations provided in the TinyverseGP framework, we perform hyperparameter optimisation of the respective representations with these two operators. Our work demonstrates that hyperparameter optimisation can lead to improvements in performance for recombination-based Cartesian Genetic Programming.
\end{abstract}


\ifarxiv\else
\fi

\ifarxiv\else 
\maketitle
\fi

\section{Introduction}\label{sec:intro}

Genetic Programming (GP), originally introduced by Cramer~\cite{Cramer1985} and Dickmanns et al.~\cite{DickmannsSW1987}, is a search paradigm for program synthesis inspired by biological evolution which evolves computer programs by means of natural selection~\cite{Koza92}. The selection pressure is implemented in the form of a fitness function that guides the search for optimised programs and a population of so-called individuals changes gradually under this selective pressure 
by means of genetic operators that introduce, 
delete or mix genetic material. The two most prominent operators are mutation and crossover. 


Cartesian Genetic Programming (CGP) is a form of GP that encodes programs as directed acyclic graphs, using a grid-based integer representation, where nodes are arranged in columns and rows and evolution selects both the graph topology and the functions computed at each node.
Because since the introduction of CGP crossover techniques have been shown to rather hinder than benefit search performance \cite{DBLP:conf/gecco/CleggWM07}, mutation, especially point mutation, has established itself as the main genetic operator for CGP \cite{DBLP:series/ncs/Miller11a}. Recent work has shown that certain forms of crossover that are adapted to CGP are indeed able to beneficially influence the search performance of CGP 
at least to some extend (see, \textit{e.g.,} \cite{DBLP:conf/eurogp/KalkreuthRD17, DBLP:conf/ppsn/Kalkreuth22}). However, recombination in CGP has been demonstrated to be disruptive when compared to recombination in linear-based GP~\cite{KochEtAl2024}. Furthermore, even if the effectiveness of crossover in CGP has been comprehensively studied quite recently~\cite{KochEtAl2024}, we believe that those studies can be enhanxced in terms of tuning the hyperparameters of the respective methods,
in order to ensure that they reach their full performance potential. 

We present an implementation and evaluation of recombination techniques for CGP that have been proposed by Kalkreuth et al. \cite{DBLP:conf/eurogp/KalkreuthRD17, DBLP:conf/ppsn/Kalkreuth22} and yielded promising results in preliminary studies. We have implemented 
the proposed crossover methods, subgraph crossover and discrete phenotypic recombination, in the TinyverseGP benchmark framework \cite{DBLP:journals/corr/abs-2504-10253}. Here, we investigate 
whether the optimisation of CGP hyperparameters included in TinyverseGP provides any benefits over manually configured hyperparameters for recombination-based CGP. To this end, 
we conducted an empirical evaluation on five different datasets from SRBench.

Our empirical results suggest that hyperparameter optimisation can lead to better results when compared to baseline configurations 
for recombination-based CGP. 
Interestingly, the positive impact of hyperparameter optimisation has been most pronounced if any type of crossover is used at all, with mutation-only CGP exhibiting almost identical performance, irrespective of whether hyperparameters were tuned or not.
Additionally, we compared the performance of CGP equipped with the subgraph crossover technique to the performance of CGP with discrete phenotypic recombination,
including as a baseline the mutation-only CGP variant (\emph{i.e.,} without crossover). 
We observed that the 
median 
performance of discrete phenotypic recombination is better than that of subgraph crossover when configured for the symbolic regression problems that we considered. 
While previous approaches have relied mostly on rules of thumb, convention, and personal experience to configure hyperparameters \cite{DBLP:series/ncs/Miller11a, KronbergerBKWA24, DBLP:conf/ppsn/Kalkreuth22}, this is not suitable to determine whether a new representation, variant or genetic operator generally has better performance than another, which motivates our work. 

The remainder of this paper is structured as follows: in Section~\ref{sec:prelim}, we briefly describe CGP, symbolic regression as our use-case, hyperparameter optimisation and the GP benchmarking framework TinyverseGP. The experiments are described in Section~\ref{sec:experiments}, followed by the presentation of the corresponding results in Section~\ref{sec:results}. Finally, in Section~\ref{sec:conclusion}, we provide concluding remarks. In addition, an overview of the implementation of the recombination operators in TinyverseGP is presented in the appendix.


\section{Preliminaries}\label{sec:prelim}

We first introduce the methods employed in this work: CGP and the crossover operators we implemented, the symbolic regression problem which we solve with it, the hyperparameter optimisation method we applied on it, and the benchmarking framework we used.

\subsection{Cartesian Genetic Programming}\label{sec:def-cgp}

CGP, as introduced by Miller and Thompson~\cite{DBLP:conf/eurogp/MillerT00}, is a graph-based variant of GP. It makes a distinction between \textit{genotype} and \textit{phenotype} regarding its encoding of computer programs. The genotype encodes the program in terms of a list of integers that describe the functions and connections of the corresponding graph. 

\begin{figure}
    \centering
    \includegraphics[scale=0.55]{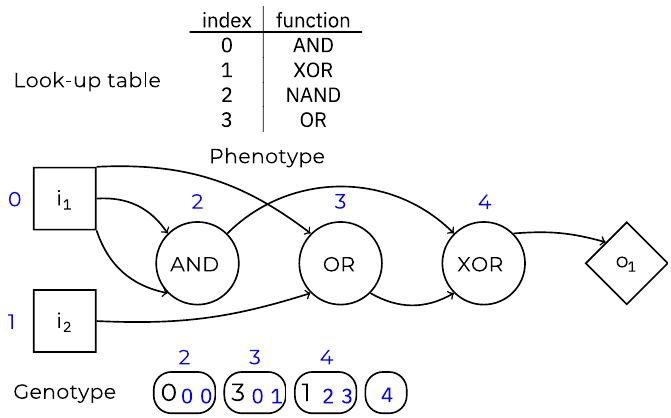}
    \caption{Exemplification of the CGP encoding.}
    \label{fig:cgp_encoding}
\end{figure}

The genotype is usually decoded in a backward manner, starting from the output nodes and following the successive input sources to determine the \textit{active nodes} of the program. The result is the actual program encoded by the genotype: the phenotype. Nodes that are not traversed in this process and their corresponding genes are called \textit{non-coding}. An illustrative example of the CGP encoding is given in Figure \ref{fig:cgp_encoding}. 

We consider two types of crossover in this work.
\textbf{Subgraph crossover} 
has been proposed by Kalkreuth et al. \cite{DBLP:conf/eurogp/KalkreuthRD17}; it takes two parents and swaps their genetic material while respecting the active nodes of each parent thus preserving some of their phenotypic properties.
\textbf{Discrete phenotypic recombination}, 
also called discrete phenotypic crossover or discrete recombination, has shown promising results when employed as genetic operator in CGP \cite{DBLP:conf/ppsn/Kalkreuth22}. The method works on two parent individuals and swaps genetic material with a focus on active nodes and creates two offspring individuals.

\subsection{Symbolic Regression} \label{sec:symbolic_regression}
According to the definition by Kronberger et al. \cite{KronbergerBKWA24} and La Cava et al. \cite{DBLP:journals/corr/LaCavaOBFVJKM21}, Symbolic Regression (SR) is a type of regression analysis which aims to find a model $$\hat{y}(\mathbf{x}) = \hat{\phi}(\mathbf{x}, \hat{\theta}): \mathbb{R}^d \mapsto \mathbb{R}$$ for features $\mathbf{x} \in \mathbb{R}^d$, where the distance between the predicted value $\hat{y}(\mathbf{x})$ and the target value $y$ is as small as possible. This is achieved by continuously altering the parameters and structure of the symbolic representation of a mathematical expression.
An illustrative example of such an expression is $10/(5 + (x - 3)^2+(y - 3)^2+(z - 3)^2 + (v - 3)^2 + (w - 3)^2)$, which is also known as Vladislavleva-4 \cite{DBLP:journals/tec/VladislavlevaSH09}, or Nguyen-11: $x^y$ \cite{DBLP:journals/gpem/NguyenNOML11}. Symbolic regression problems are often tackled using GP, where different types of error between the predicted value and the target value, \textit{e.g.,} mean squared error (MSE), are used to assess the fitness of the model. 

\subsection{Hyperparameter Optimisation} \label{sec:bayesian_optimisation}

Hyperparameter optimisation is the problem of finding a hyperparameter configuration for a machine learning algorithm that leads to optimal performance of that algorithm on the task at hand. It allows to automatically tailor a method, here CGP, to the problem instances of interest, here a symbolic regression task. 
Hyperparameter optimisation has been shown to be crucial for achieving strong performance across a wide range of algorithms and application domains~\cite{LavEtAl2006,SnoEtAl2012,BergEtAl2012}.
In this work, we use SMAC3 \cite{DBLP:journals/jmlr/LindauerEFBDBRS22}, a state-of-the-art 
general-purpose algorithm configurator that uses \textit{Bayesian optimisation} \cite{Garnett23, Frazier18} to efficiently search for optimal configurations, balancing exploration and exploitation within a given search space.  
\subsection{TinyverseGP Benchmarking Framework}
\label{sec:tinyversegp}
TinyverseGP \cite{DBLP:journals/corr/abs-2504-10253} is a Python benchmarking library for GP that incorporates several representations of GP such as CGP, tree-based GP, linear-based GP, and grammatical evolution. It provides a unified framework in which the performance of GP representations (using different measures) can be tested and compared to each other, using problems from several problem domains, including symbolic regression (one of the most prominent problem domains in GP \cite{DBLP:journals/gpem/WhiteMCMGKJOL13}) as well as logic synthesis and policy learning. 
In particular, TinyverseGP contains interfaces to problems from SRBench \cite{DBLP:journals/corr/LaCavaOBFVJKM21}, a benchmarking suite for symbolic regression problems, which is our use-case of interest in this work, 
and an interface to SMAC3 \cite{DBLP:journals/jmlr/LindauerEFBDBRS22}, which enables hyperparameter optimisation.

\sloppy
Concretely, in this work, the evaluated operators have been implemented in TinyverseGP and required minor adjustments.
The pseudocode of the functions \texttt{subgraph\_crossover} and \texttt{discrete\_recombination} can be found in Appendix~\ref{sec:pseudocode}, and implementation details and necessary modifications are described in Appendix~\ref{sec:implementation}.

\section{Experiments}
\label{sec:experiments}

This section provides details 
on the experiments we have conducted in order to answer the following research questions:\\
\textbf{RQ 1:} What is the impact of hyperparameter optimisation on the performance of recombination-based CGP?\\
\textbf{RQ 2:} How does the performance of CGP with subgraph crossover compare to that of CGP with discrete phenotypic recombination?

To answer these questions, we performed hyperparameter optimisation using SMAC3 on black-box symbolic regression problems. 
We used five datasets from the Penn Machine Learning Benchmarks (PMLB) collection of datasets \cite{OlsonLOUM17, RomanoLLGGCRHFM21}. 
We then evaluated recombination-based CGP using both manually chosen baseline hyperparameter configurations and the hyperparameter configurations optimised by SMAC3.

\subsection{Setup of Experiments}

A canonical evolutionary algorithm equipped with tournament selection was adapted to run recombination-based CGP. The algorithm received hyperparameters such as population size, size of the tournament, crossover and mutation probabilities, which defined the hyperparameter-space considered by SMAC3. The HPO scenario for SMAC3 used for the experiments is available in our repository\footnote{\url{https://github.com/GPBench/TinyverseGP/tree/recombination-based-cgp/src/hpo/hpo_cgp_sr.py}}.

For each dataset, crossover operator and hyperparameter values, we ran CGP 30 times, in order to obtain a distribution of values and report statistics on this distribution as final performance.

We measured the fitness of a CGP run on a symbolic regression problem as the error between the predicted value of the evolved expression and the target value of the expression; specifically, we used MSE as a minimisation objective.

\subsection{Datasets}


We used five datasets from the Penn Machine Learning Benchmarks collection (PMLB)~\cite{OlsonLOUM17, RomanoLLGGCRHFM21}. The characteristics, such as number of variables and observations, are shown in \autoref{tab:datasets}. The selected datasets are part of SRBench~\cite{DBLP:journals/corr/LaCavaOBFVJKM21}. Following~\cite{AldeiaEtAl2025}, we chose small datasets with less than $500$ observations 
and real-valued target functions, \textit{i.e.,} we excluded integer-valued targets and large datasets.


\begin{table}
  \centering
  \small
  \scalebox{0.8}{
    \begin{tabular}{lrr}
    \toprule
    \textbf{Dataset} & \textbf{\# Variables} & \textbf{\# Observations}\\
    \midrule
    \href{https://epistasislab.github.io/pmlb/profile/192_vineyard.html\#top}{\texttt{192\_vineyard}} & 3 & 52 \\
    \midrule
    \href{https://epistasislab.github.io/pmlb/profile/210_cloud.html\#top}{\texttt{210\_cloud}} & 6 & 108  \\
    \midrule
    \href{https://epistasislab.github.io/pmlb/profile/579_fri_c0_250_5.html\#top} {\texttt{579\_fri\_c0\_250\_5}} & 6 & 250 \\
    \midrule
    \href{https://epistasislab.github.io/pmlb/profile/650_fri_c0_500_50.html\#top} {\texttt{650\_fri\_c0\_500\_50}} & 20 & 500  \\
    \midrule
    \href{https://epistasislab.github.io/pmlb/profile/678_visualizing_environmental.html} {\texttt{678\_visualizing\_environmental}} & 4 & 11  \\
    \bottomrule
    \end{tabular}
    }
    \caption{Datasets from SRBench used in our experiments and their characteristics.}
    \label{tab:datasets}
\end{table}

\subsection{Baseline Runs}

As a baseline, we performed two separate sets of CGP runs with manually configured hyperparameters. The manually chosen values for the hyperparameters can be seen in Table \ref{tab:base_hyperparams}. 
The sets differ in the number of nodes, which is set to $100$ and $200$, respectively, with all other hyperparameters fixed. They are based on  studies conducted in the context of the introduction of the subgraph crossover and discrete phenotypic recombination operators, where the new operators have also been tested on problems from the domain of symbolic regression \cite{DBLP:conf/eurogp/KalkreuthRD17, DBLP:conf/ijcci/Kalkreuth20, DBLP:conf/ppsn/Kalkreuth22}. A population size of $50$ has been determined in the meta-evolution process using the Java Evolutionary Computation Research (ECJ) tools \cite{Luke98, DBLP:conf/gecco/ScottL19} for all symbolic regression problems for discrete recombination \cite{DBLP:conf/ppsn/Kalkreuth22} and for many of the symbolic regression problems in the study on subgraph crossover \cite{DBLP:conf/ijcci/Kalkreuth20}.

We used the same fixed pseudo-random number seeds for the baseline runs and for the configured runs, to obtain the same training and testing sets for a fair comparison and to ensure reproducibility. We evolved the model on the training set and then took the best individual to calculate 
for further evaluation on the held-out testing set.

\subsection{Hyperparameter Optimisation Runs}

We optimised the hyperparameters listed in \autoref{tab:base_hyperparams}. For each dataset, we split the data into a training and testing set using a fixed seed, 
where the training and testing sets make up $75\%$ and $25\%$ of the dataset, respectively. We tuned the hyperparameters of CGP using SMAC3 on the training set via $5$-fold cross-validation with a budget of $200$ trials.
For reproducibility, the $5$-fold cross-validation was seeded with a fixed value for all runs. 
We optimised for the mean of the fitness values of the best individual on the validation set of each iteration of the $5$-fold cross-validation.
For each dataset, we performed $10$ independent runs of the hyperparameter optimisation procedure times with random seeds, each yielding one optimised hyperparameter configuration.
We then performed $30$ runs of of each of these $10$ optimised hyperparameter configurations for each crossover approach, with seeds in $\{1, 2,  \dots, 10\}$. 
For all runs, a seed with value 42 was used for the CGP and for the split of the dataset.

\begin{table}
  \centering
  \scalebox{0.8}{
    \small
    \begin{tabular}{ll}
    \toprule
    \textbf{Hyperparameter} & \textbf{Value}\\
    \midrule
    population\_size & 50\\
    \midrule
    levels\_back & 100\\
    \midrule
    mutation\_rate & 0.1\\
    \midrule
    cx\_rate & 0.7\\
    \midrule
    tournament\_size & 4\\
    \midrule
    num\_function\_nodes & 100 or 200\\
    \bottomrule
\end{tabular}
}
    \caption{Baseline hyperparameters for the evaluation runs.}
    \label{tab:base_hyperparams}
\end{table}

\section{Results}
\label{sec:results}

\begin{figure}
\begin{subfigure}{0.47\textwidth}
    \includegraphics[width=0.95\linewidth]{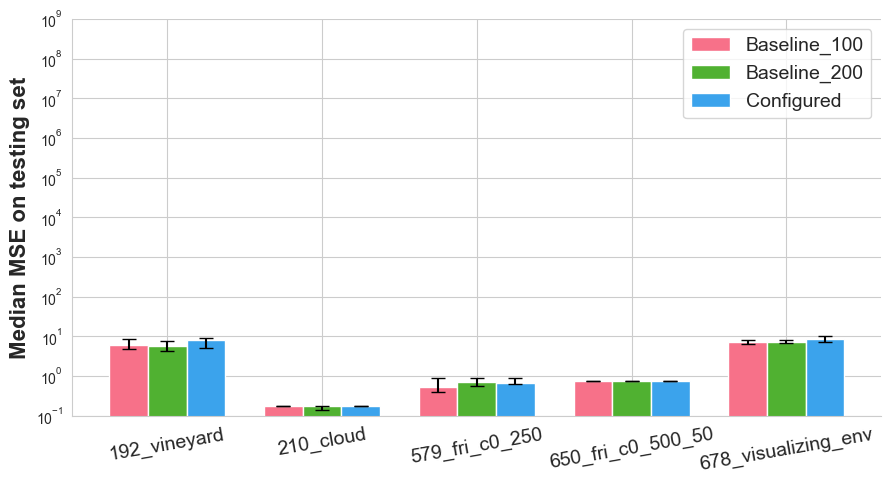}
    \caption{Mutation-only}
    \label{fig:mutation}
\end{subfigure}

\begin{subfigure}{0.47\textwidth}
    \includegraphics[width=0.95\linewidth]{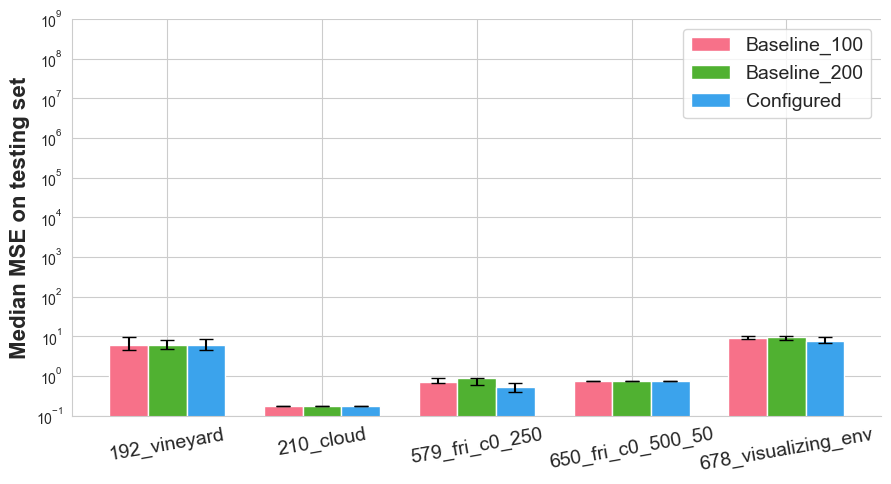}
    \caption{Discrete recombination}
    \label{fig:discrete_recombination}
\end{subfigure}

\begin{subfigure}{0.47\textwidth}
    \includegraphics[width=0.95\linewidth]{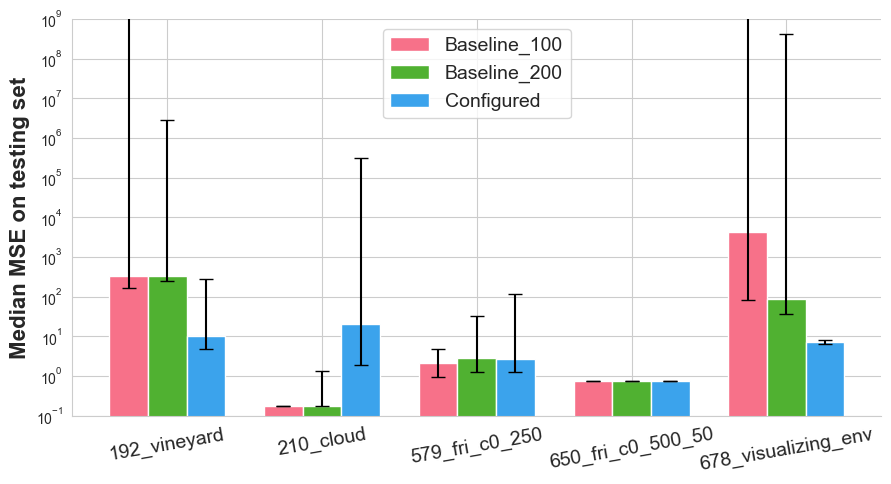}
    \caption{Subgraph crossover}
    \label{fig:subgraph_crossover}
\end{subfigure}
\caption{Median fitness of the best found configuration against our baselines. Error bar shows the first and third quartiles. The truncated error bars from \texttt{192\_vineyard} and \texttt{678\_visualizing\_env} 
are reaching 
$1.18 \cdot 10^{11}$ and $1.55 \cdot 10^{19}$, respectively.}
\label{fig:configuration}
\end{figure}

To answer \textbf{RQ 1}, Figure~\ref{fig:configuration} shows the median fitness (in terms of MSE) of the best hyperparameter configuration found by SMAC3 against the two baseline configurations (100 and 200 nodes) across all five datasets. 
The error bars indicate the first and third quartiles of the underlying distribution.
For both discrete recombination and subgraph crossover, the optimised configuration consistently matches or outperforms at least one of the baselines across all datasets. 
The benefit of hyperparameter optimisation is particularly visible on datasets such as \texttt{579\_fri\_c0\_250} for discrete recombination and \texttt{678\_visualizing\_environmental} for subgraph crossover, where the 
optimised configurations achieve substantially lower MSE values than either baseline. These results suggest that hyperparameter optimisation can meaningfully improve the performance of recombination-based CGP beyond what is achievable with manually chosen configurations.

\begin{table}[]
\small
\scalebox{0.8}{
    \begin{tabular}{lrrrrrr}
\toprule
\textbf{Dataset} & \multicolumn{2}{c}{\makecell{\textbf{Discrete}\\\textbf{Recombination}}} & \multicolumn{2}{c}{\makecell{\textbf{Subgraph}\\\textbf{Crossover}}} & \multicolumn{2}{c}{\makecell{\textbf{Mutation-only}}}\\
& median & \makecell{\footnotesize Q1 \\[-3pt]\footnotesize  Q3}& median & \makecell{\footnotesize Q1 \\[-3pt]\footnotesize  Q3}& median & \makecell{\footnotesize Q1 \\[-3pt]\footnotesize  Q3}\\
\midrule
\texttt{192\_vineyard} & \underline{\it 6.20} & \makecell[r]{\footnotesize 4.62 \\[-3pt]\footnotesize  8.78} & 10.55 & \makecell[r]{\footnotesize 4.82 \\[-3pt]\footnotesize  284.83} &  \it 8.31 & \makecell{\footnotesize 5.06 \\[-3pt]\footnotesize  9.03} \\
\texttt{210\_cloud} & \underline{\it 0.18} & \makecell[r]{\footnotesize 0.18 \\[-3pt]\footnotesize  0.18} & 20.34 & \makecell[r]{\footnotesize 1.89 \\[-3pt]\footnotesize  314882.07} & \it 0.18 & \makecell{\footnotesize 0.18 \\[-3pt]\footnotesize  0.18} \\
\texttt{579\_fri\_c0\_250} & \underline{\it 0.54} & \makecell[r]{\footnotesize 0.40 \\[-3pt]\footnotesize  0.68} & 2.62 & \makecell[r]{\footnotesize 1.28 \\[-3pt]\footnotesize  114.59} & 0.68 & \makecell{\footnotesize 0.63 \\[-3pt]\footnotesize  0.88} \\
\texttt{650\_fri\_c0\_500\_50} & \underline{\it 0.74} & \makecell[r]{\footnotesize 0.74 \\[-3pt]\footnotesize  0.74} & \it 0.74 & \makecell[r]{\footnotesize 0.74 \\[-3pt]\footnotesize  0.74} & \it 0.74 & \makecell{\footnotesize 0.74 \\[-3pt]\footnotesize  0.74}\\
\texttt{678\_visualizing\_env} & \it 7.69 & \makecell[r]{\footnotesize 6.95 \\[-3pt]\footnotesize  9.42} & \underline{\it 7.26} & \makecell[r]{\footnotesize 6.60 \\[-3pt]\footnotesize  8.23} & 8.45 & \makecell{\footnotesize 7.44 \\[-3pt]\footnotesize  10.14}\\
\bottomrule
\end{tabular}
}
    \caption{Median performance (in terms of MSE) of the configured CGP models on each dataset (lower is better) with first and third quartile. The best median is underlined, and the medians statistically tied with the best, according to a Mann-Whitney U-test with a significance level of $0.05$ are in italic. 
    }
    \label{tab:operatorresult}
\end{table}


To answer \textbf{RQ 2}, Table~\ref{tab:operatorresult} shows the median MSE of the best configurations found for each operator across our five datasets and highlights the median statistically tied with the best configuration according to a
Mann-Whitney U-test with a significance level of 0.05. 
Discrete recombination achieves lower MSE on four out of the five datasets, with subgraph crossover only performing comparably on the \texttt{678\_visualizing\_environmental} dataset. 
Overall, discrete phenotypic recombination demonstrates stronger median performance on the symbolic regression problems considered here. In comparison to mutation-only CGP, the considered recombination-based approaches overall achieved better or at least equal median performance, while discrete recombination showed more robust performance. 
Additionally, the variability in performance of CGP is especially pronounced with subgraph crossover as an operator (Figure~\ref{fig:subgraph_crossover}), as the scale of achieved median performance values with subgraph crossover are up to two order of magnitude larger than the one with discrete recombination.
(Figure~\ref{fig:discrete_recombination}), with the third quartile being more than ten orders of magnitude larger on the \texttt{192\_vineyard} and \texttt{678\_visualizing\_environmental} datasets. 
However, the configured mutation-only CGP, does not to differ much from the mutation-only baselines (Figure~\ref{fig:mutation}). 

\section{Conclusions and Future Work}\label{sec:conclusion}

We presented a first step towards improved evaluation of recombination-based CGP to facilitate fairer comparisons between operators and algorithms. We have reported results for recombination-based and mutation-only CGP that have been obtained by using a pipeline implemented in TinyverseGP, which enables hyperparameter optimisation using SMAC3 with 5-fold cross-validation and subsequent evaluation of these optimised hyperparameter settings. To ensure fairer comparison of genetic operators, our results demonstrate that hyperparameter optimisation should be used to find suitable configurations for the evaluated operators, and the comparison can be made based on the configurations with the best performance of each operator or method. 

In future work, we plan to extend the proposed approach to other recombination operators from the literature,
in order to evaluate CGP in the context of a larger study built upon best practices in benchmarking randomised search heuristics~\cite{bartzbeielstein2020benchmarkingoptimizationbestpractice}. 
We also intend to carry out fitness landscape analysis to learn more about the optimisation conditions in GP model tuning. Since for some of the datasets we tested, the best found configuration performed worse than our baselines, we would naturally like to better understand the reasons and corresponding challenges. 


\ifarxiv\else\balance\fi
\bibliographystyle{ACM-Reference-Format}
\bibliography{gecco26} 

\clearpage

\appendix

\onecolumn

\section{Pseudocode}
\label{sec:pseudocode}

\begin{algorithm}[H]
\caption{Subgraph crossover}
\label{alg:subgraph_crossover}
\KwData{Genome of the first parent $P_1$, genome of the second parent $P_2$}
\KwResult{Offspring genome $O$}
\Comment{Store the active nodes of each parent in arrays}
$M_1 \leftarrow \text{ActiveNodes}(P_1)$\;
$M_2 \leftarrow \text{ActiveNodes}(P_2)$\;
\Comment{Choose crossover point for each parent}
$C_{P_1} \leftarrow \text{RandomChoice}(M_1)$\;
$C_{P_2} \leftarrow \text{RandomChoice}(M_2)$\;
\Comment{Choose general crossover point to be the minimum of the two crossover points $C_{P_1}$ and $C_{P_2}$}
$C_P \leftarrow \text{Min}(C_{P_1}, C_{P_2})$\;
\Comment{Copy the genomes of the parents}
$O\leftarrow \text{CopyGenome}(P_1,\text{start}, C_P)$\;
$O\leftarrow \text{CopyGenome}(P_2,C_P + 1, \text{end})$\;
\Comment{Perform neighbourhood connect}
\If{$\text{NumberOfOutputs} = 1$}{
    \Comment{Connect first connection gene behind $C_P$ to last active node of the subgraph in front of $C_P$}
    $O[\text{NodePosition}(\text{FirstActive + 1})] \leftarrow \text{NodePosition}(\text{LastActive})$\;
}
\Comment{Perform random active connect}
\For{gene in genome of the offspring behind $C_P$}{
    \If{gene is connection gene}{
        gene $\leftarrow$ RandomChoice(Inputs, genes in columns in front of current column)\;
    }
}
\If{$NumberOfOutputs > 1$}{
    \For{gene in output genes of the offspring}{
        gene $\leftarrow$ RandomChoice(Inputs, function nodes)\;
    }
}
\end{algorithm}

\begin{algorithm}[H]
\caption{Discrete phenotypic recombination (adjusted from \cite{DBLP:conf/ppsn/Kalkreuth22})}
\label{alg:discrete_recombination}
\KwData{Genomes of the parents $G_1, G_2$}
\KwResult{Genomes of the offspring $\tilde{G}_1, \tilde{G}_1$}
\Comment{Lists storing the active node numbers of each parent}
$M_1 \gets \text{ActiveNodes}(G_1)$\;
$M_2 \gets \text{ActiveNodes}(G_2)$\;
\Comment{Determine the min and max number of active function nodes}
$\text{min} \gets \text{Min}(\text{len}(M_1), \text{len}(M_2))$\;
$\text{max} \gets \text{Max}(\text{len}(M_1), \text{len}(M_2))$\;
$i \gets 0$\;
\While{$i < 0$}{
    \Comment{Randomly choose to either swap genes or not}
    \If{$\text{RandomBoolean()}= \text{true}$}{
        \eIf{$i = \text{min} - 1 \textbf{ and } \text{len}(M_1) \neq \text{len}(M_2)$}{
            \Comment{Boundary extension}
            $r \gets \text{RandomInteger}(0, \text{max} - i)$\;
            \eIf{$\text{len}(M_1) < \text{len}(M_2)$}{
                $n_1 \gets N_1[i]$\;
                $n_2 \gets N_2[i + r]$\;
            }{
                $n_1 \gets N_1[i + r]$\;
                $n_2 \gets N_2[i]$\;
            }
        }{
            \Comment{No boundary extension}
            $n_1 \gets N_1[i]$\;
            $n_2 \gets N_2[i]$\;
        }
        $p_1 \gets \text{NodePosition}(n_1)$\;
        $p_2 \gets \text{NodePosition}(n_2)$\;
        $\tilde{G}_1, \tilde{G}_1 \gets \text{SwapGenes}\;(G_1,G_2,p_1,p_2)$\;
    }
    $i \gets i + 1$\;
} 
\end{algorithm}


\section{Implementation details}
\label{sec:implementation}


The operators evaluated in this work have been implemented in the TinyverseGP framework and required minor adjustments described below. The code will be included in the benchmark for future use by the community. The pseudocode of the functions \texttt{subgraph\_crossover} and \texttt{discrete\_recombination} can be found in the appendix. The extended version of \texttt{tiny\_cgp} is provided in our repository\footnote{\url{https://github.com/GPBench/TinyverseGP/tree/recombination-based-cgp/src/gp/tiny_cgp.py}}.

\textbf{Subgraph crossover: }For the subgraph crossover operator, we made some adjustments to the original description by Kalkreuth et al. \cite{DBLP:conf/eurogp/KalkreuthRD17} to fit the broader module of \texttt{tiny\_cgp}. The implemented method takes the genomes \texttt{genome1} and \texttt{genome2} of parents $P_1$ and $P_2$ as input (as opposed to the parent individuals as a whole). The genomes are represented as lists of type \texttt{int} and the function returns an object called \texttt{CGPIndividual} that contains the genotype of the generated offspring. The first step is analogous to the original definition and assigns the node numbers of the active nodes of parents $P_1$ and $P_2$ to lists \texttt{M\_1} and \texttt{M\_2}, respectively.

\textbf{Discrete recombination:}
The implementation of discrete phenotypic recombination largely follows the instructions that are lined out in the pseudocode given in the original paper by Kalkreuth \cite{DBLP:conf/ppsn/Kalkreuth22}, with some adjustments to fit the \texttt{tiny\_cgp} module. Similar to the subgraph crossover method and the original algorithm description, discrete recombination 
takes as input the genomes of the parents in the form of lists of integers, but returns a tuple of two CGP genotypes for the two offspring. 

\end{document}

\begin{definition}[Cartesian Genetic Programming \cite{DBLP:phd/dnb/Kalkreuth21}]
    Cartesian Genetic Programming is a form of Genetic Programming, which offers a graph-based representation model for Genetic Programming based on a rectangular grid or row of nodes.
\end{definition}

\begin{definition}[Cartesian Genetic Program \cite{DBLP:phd/dnb/Kalkreuth21}]
    A cartesian genetic program $P$ is an element of $N_i \times N_f \times N_o \times F$:

    \begin{itemize}
        \item $N_i$ is a finite non-empty set of input nodes,
        \item $N_f$ is a finite set of function nodes,
        \item $N_o$ is a finite non-empty set of output nodes.
        \item $F$ is a finite non-empty set of functions.
    \end{itemize}
\end{definition}